\documentclass[10pt,twocolumn,letterpaper]{article}
\usepackage{cvpr}
\usepackage{graphicx}
\usepackage{amsmath}
\usepackage{arydshln}
\usepackage{booktabs}
\usepackage{times}
\usepackage{epsfig}
\usepackage{amssymb}
\usepackage{dsfont}
\usepackage{color}
\usepackage{xcolor}
\usepackage{microtype} 
\usepackage{pifont}
\usepackage{float}
\usepackage[pagebackref,breaklinks,colorlinks]{hyperref}
\usepackage[capitalize]{cleveref}
\crefname{section}{Sec.}{Secs.}
\Crefname{section}{Section}{Sections}
\Crefname{table}{Table}{Tables}
\crefname{table}{Tab.}{Tabs.}

\newcommand{\mcdropout}{MC Dropout\xspace}
\newcommand{\obsnet}{ObsNet\xspace}

\newcommand{\victor}[1]{\textcolor{black}{#1}}

\newcommand{\vx}{\boldsymbol{x}}

\begin{document}


\title{\vspace{-2cm}Instance-Aware Observer Network for Out-of-Distribution Object Segmentation}

\author{
Victor Besnier$^{1,3,4}$
\and 
Andrei Bursuc$^{2}$
\and 
David Picard$^{3}$ 
\and 
Alexandre Briot$^{1}$
\and 
1. Valeo, Créteil, France
\and 
2. Valeo.ai, Paris, France
\and 
3. LIGM, Ecole des Ponts, Univ Gustave Eiffel, CNRS, Marne-la-Vallée, France
\and
4. ETIS UMR8051, CY Université, ENSEA, CNRS, Cergy France
}

\maketitle

\begin{abstract}
Recent works on predictive uncertainty estimation have shown promising results on Out-Of-Distribution (OOD) detection for semantic segmentation. However, these methods struggle to precisely locate the point of interest in the image, i.e, the anomaly. This limitation is due to the difficulty of fine-grained prediction at the pixel level. To address this issue, we build upon the recent ObsNet approach by providing object instance knowledge to the observer. We extend ObsNet by harnessing an instance-wise mask prediction. We use an additional, class agnostic, object detector to filter and aggregate observer predictions. Finally, we predict an unique anomaly score for each instance in the image. We show that our proposed method accurately disentangles in-distribution objects from OOD objects on three datasets. 
\end{abstract}

\vspace{-0.45cm}
\section{Introduction}\label{sec:introduction}

Lately, an ever increasing number of safety-critical systems, such as autonomous driving, 
are looking into leveraging Deep Neural Networks (DNNs) for the perception of the environment and as well as for subsequent decisions. Despite some success, DNNs still remain unreliable for real world deployment and often over-confident even they are incorrect on both In-Distribution~\cite{guo_2017} and Out-Of-Distribution (OOD) data~\cite{nguyen2015deep,hein2019relu}. In this work, we aim at detecting OOD objects for 2D object segmentation. In this context, we consider as OOD the objects belonging to a class that is unknown by the perception system, i.e., a class that is not defined nor present in the training data.

Most methods dealing with \textit{unknown-unknown} are based on ensembles\cite{Lakshminarayanan_2017,maddox2019simple,franchi2019tradi,mehrtash2020pep}, pseudo-ensembles~\cite{KendallGal_2017,wen2020batchensemble,franchi2020encoding,durasov2021masksembles}, or deterministic approaches for computing uncertainty~\cite{hendrycks17baseline,postels2019sampling, van2020uncertainty,liu2020simpleSNGP}. However, most of them cannot simultaneously satisfy the real world requirements of high performance and real-time inference, typically trading one for the other. Recent works inspired by practices from system validation and monitoring, advance two-stage strategies to detect anomalies in semantic segmentation~\cite{besnier2021learning,besnier2021triggering}. An Observer Network is trained to analyze and predict the confidence of a main perception network. Observer-based approaches have been shown to find a good balance between accuracy and computational efficiency~\cite{besnier2021learning,besnier2021triggering}. In this work we build on top on observer-based approaches to leverage their properties.

We argue that pixel-wise error map (as shown in ~\cite{besnier2021learning,besnier2021triggering,KendallGal_2017, corbiere2019addressing}) by itself is sub-optimal for anomaly detection in segmentation because these maps lack clarity. Due to the difficulty of fine-grained prediction, most boundaries between two classes as well as small or distant objects are considered uncertain. Therefore, the focus of interest in the image, i.e. the OOD object, is drowned into this noise. The resulting error map does not provide precisely delimited spatial information: we know there is an error on the image but we struggle to accurately locate the corresponding OOD object. In other words, while we can get uncertainty estimates and predictions at pixel-level, extending them to objects is not obvious and we cannot find automatically objects far from the training distribution. As an example, an image depicting a crowd of pedestrian with lot of boundaries has, on average, higher anomaly score than an image with only one OOD object in the middle of a road.

\begin{figure}
\renewcommand{\captionfont}{\small}
\renewcommand{\captionlabelfont}{\bf}
    \centering
    \includegraphics[width=\linewidth]{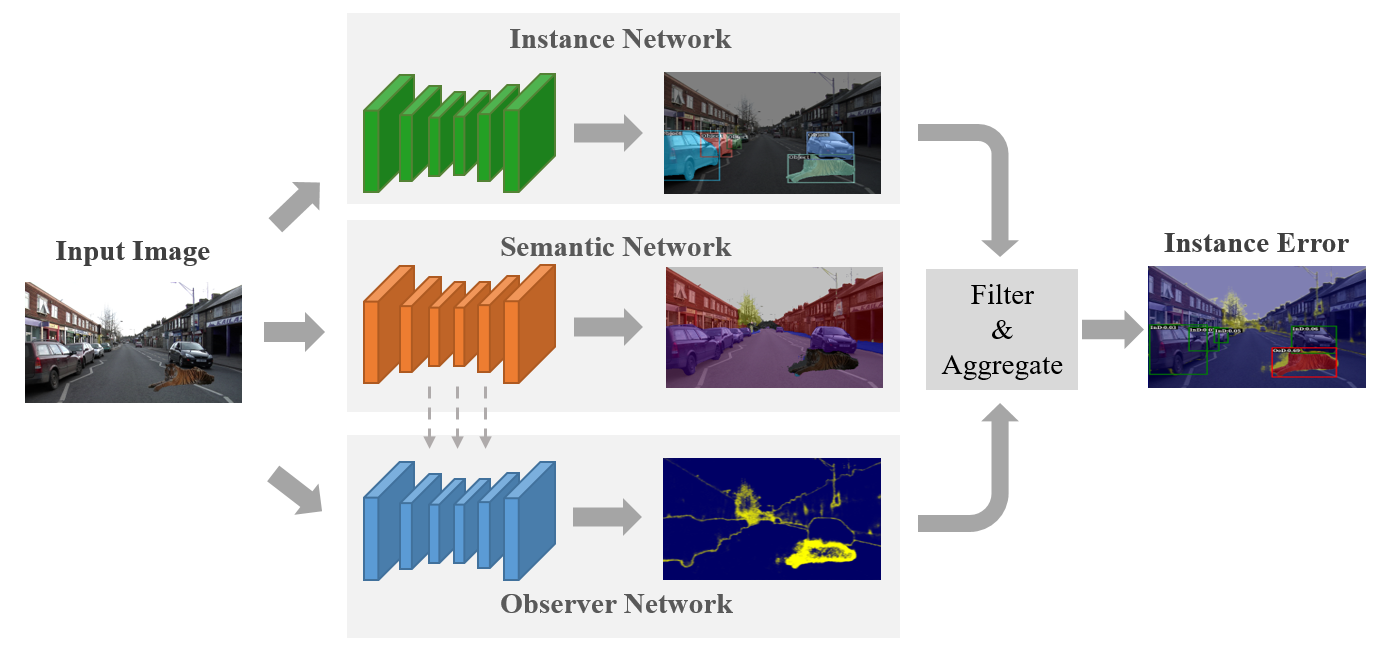}
    \caption{\textbf{Overview of our instance aware pipeline.} The image is fed into the instance, the semantic and the observer network (from top to bottom). On the middle, the \obsnet prediction is filter by the class-agnostic instance prediction and the remaining error is then aggregate object-wise. Finally, in the right of the figure, we show in red, the objects far from the training distribution.}
    \label{fig:architecture_instance}
\vspace{-0.45cm}
\end{figure}

In this paper, we propose to reduce the granularity of the task in order to improve the relevance of the error map. To this end, we use a class agnostic, instance segmentation network. With this additional prediction, we first filter background errors, and then aggregate uncertainty in an instance-aware manner \autoref{fig:architecture_instance}. Ultimately, we only want to highlight object instances with high errors. With this pragmatic and practical solution we can sort objects by anomaly score and then discard all objects close to the training distribution and keep those that are far from the training distribution.

\begin{figure*}
\renewcommand{\captionfont}{\small}
\renewcommand{\captionlabelfont}{\bf}
\centering
\includegraphics[width=\linewidth]{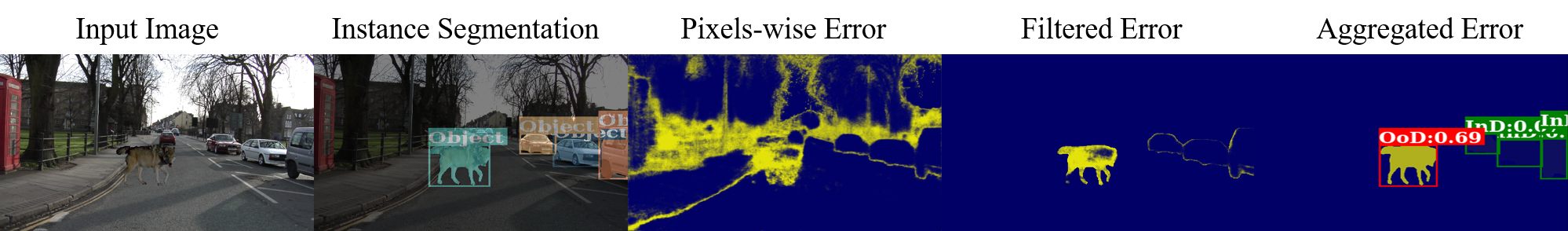}
\caption{\victor{\textbf{Flows of the image processing.} From the input image (left), we compute the pixels-wise uncertainty and the object detection masks. Then we filter the uncertainty in the area of object only, and finally aggregate the score in an instance aware manner. We can see that the OOD object is well detected while in-distribution objects with low anomaly score and background errors are erased (right).}}
\label{fig:filter_uncertainty}
\vspace{-0.3cm}
\end{figure*}
\section{Related Work}\label{section:related_work}
The widespread DNN adoption in this field has led to a fresh wave of approaches to improve OOD detection by input reconstruction \cite{schlegl2017unsupervised, baur2018deep, lis2019detecting, xia2020synthesize}, predictive uncertainty \cite{Gal2016Dropout,KendallGal_2017, malinin2018}, ensembles \cite{Lakshminarayanan_2017,franchi2019tradi,durasov2021masksembles}, adversarial attacks \cite{Liang2018, lee2018simple}, using a void or background class \cite{ren2015faster, liu2016ssd} or dataset \cite{bevandic2019simultaneous, hendrycks2018deep, malinin2018}, \etc., to name just a few. 

\noindent\textbf{Bayesian approaches and ensembles.} BNNs~\cite{neal2012bayesian,Blundell2015} can capture predictive uncertainty from learned distributions over network weights, but don't scale well~\cite{dusenberry2020efficient} and approximate solutions are preferred in practice. Deep Ensembles (DE)~\cite{Lakshminarayanan_2017} is a highly effective, yet costly approach, that trains an ensemble of DNNs with different initialization seeds. Efficient or pseudo-ensemble approaches are a pragmatic alternative to DE that bypass training of multiple networks and generate predictions from different random subsets of neurons~\cite{Gal2016, srivastava2014dropout,durasov2021masksembles} or from networks sampled from approximate weight distributions~\cite{maddox2019simple,franchi2019tradi,mehrtash2020pep, wen2020batchensemble,franchi2020encoding}. However they all require multiple forward passes and/or storage of additional networks in memory.

\noindent\textbf{Learning to predict errors.} Inspired by early approaches from model calibration literature~\cite{platt1999probabilistic, zadrozny2001obtaining, zadrozny2002transforming, naeini2015obtaining, naeini2016binary}, a number of methods propose endowing the task network with an error prediction branch allowing self-assessment of predictive performance. This branch can be trained jointly with the main network~\cite{devries2018learning, yoo2019learning}, however better learning stability and results are achieved with two-stage sequential training~\cite{corbiere2019addressing, hecker2018failure,besnier2021learning,Samson_2019_ICCV}.

\noindent\textbf{OOD in object detection.} 
Compared to classification and segmentation, OOD identification for 2D object detection is more challenging (mix of classification and regression predictions) and is less explored. Here, ensembles and pseudo-ensembles are often used for predictive uncertainty~\cite{miller2019evaluating,haussmann2020scalable, azevedo2020stochastic,rahman2020online}. Spatial uncertainty can be computed with slight changes in popular architectures and losses \cite{Choi_2019_ICCV,Harakeh2021EstimatingAE,choi2021active}. Other works leverage sampling-free approaches for both spatial and class uncertainty for real-time applications~\cite{Gasperini_2022}. OOD detection itself has started to be addressed only recently by generating outliers in the feature space~\cite{du2022towards} or by ``seeing'' unknown objects from videos in the wild~\cite{du2022stud}.

\section{Proposed Method}\label{section:method}

\subsection{Observer Networks}
Our work builds upon \obsnet \cite{besnier2021learning,besnier2021triggering} that we briefly describe here. Observer networks are a two-stage method to detect pixels-wise errors and OOD. \cite{besnier2021triggering} designed two principles to train efficiently an auxiliary network. They improve the architecture by decoupling the OOD detector from the segmentation branch and by observing the whole network via residual connections. Secondly, they generate blind spots in the segmentation network with \textit{local adversarial attacks} ($LAA$) at a random location of the image, mimicking an OOD object. \obsnet ($Obs$) outputs a pixels-wise error map corresponding to the probability that the semantic segmentation network ($Seg$) fails to predict the correct class $y$:

\begin{equation}
    Obs(\vx, Seg_r(\vx))\approx Pr[ Seg(\vx) \neq y ],
\end{equation}
where $x$ is the input image and $Seg_r$ the skip connections from intermediate feature maps of segmentation network $Seg$.

\subsection{Instance Anomaly Detection}
To this end, we upgrade the semantic segmentation framework with instance hints. We use an additional class agnostic instance segmentation prediction. This detector ($Det$) produces a binary mask by mapping each object in the image.

Then, the idea is to separate the observer's prediction map into two categories. The background (classes of \emph{stuff}) and the instance (classes of \emph{things}) in the same way as the panoptic segmentation. Background errors correspond to global ambiguities in the scene at different scales: error at the decision boundary between two classes, prediction error between the road and the sidewalk or complexity of the leaves of a tree. In contrast, an instance error corresponds to an object far from the train distribution. 

\subsection{Uncertainty Aggregation and Filtering}
In order to obtain a unique error score for each instance (similar to the well-known objectness score in object detection), we aggregate the per-pixel uncertainty within the predicted object mask to a unique value. In practice, given an image $\vx \in \mathbb{R}^{3 \times H \times W}$, we predict for each detected object $o_i$ an anomaly score $a_{i} \in \mathbb{R}$:

\begin{equation}\label{eq:anomalyscore}
    a_{i} = \frac{1}{M} \sum_{h=0}^{H}\sum_{w=0}^{W}{u^{(h,w)} \odot m^{(h,w)}_{i}},
\end{equation}
where $u = Obs(\vx, Seg_r(\vx)) \in \mathbb{R}^{H \times W}$ is the pixel-wise error map of \obsnet; $m_{i} \in \mathbb{R}^{H \times W}$ is the binary mask of an instance $o_i$ in the set of the detector prediction $Det(\vx)=\{m_i\}$; $M=\sum_{h,w=0}^{H \times W}{m_i}$ the area of the instance $o_i$; and $\odot$ is the element-wise product. We also filter predicted instance masks $m_i$ by size, in order to remove very small detected objects ($<16^2$ pixels) in the image. 

This strategy shows several benefits. We can discover instances in the dataset that do not match with the training distribution, useful for active learning or object discovery. We can also localize where the anomaly is in the image, which is a primary requirement for safety-critical applications such as autonomous driving. In \autoref{fig:filter_uncertainty}, we show that our framework is able to detect several instances in the images, and the \obsnet succeeds in discriminating in-distribution objects from out-of-distribution ones. 

\section{Experiments}\label{section:results}
\begin{figure}
\renewcommand{\captionfont}{\small}
\renewcommand{\captionlabelfont}{\bf}
\centering
\includegraphics[width=\linewidth]{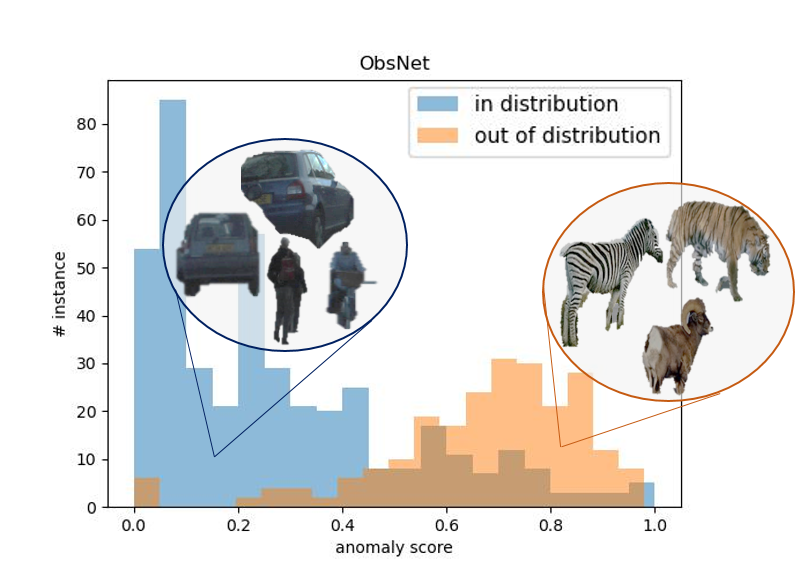}
\caption{\textbf{Histogram on CamVid OOD}. Anomaly score from obsnet and detection from mask RcNN trained on the pan-distribution. We show here some examples of well-detected objects and predicted as in distribution in blue (left). While objects detected with high anomaly score (right) are considered as OOD in orange.}
\label{fig:histogram}
\vspace{-0.3cm}
\end{figure}

\begin{figure*}
\renewcommand{\captionfont}{\small}
\renewcommand{\captionlabelfont}{\bf}
\centering
\includegraphics[width=\linewidth, height=5.5cm]{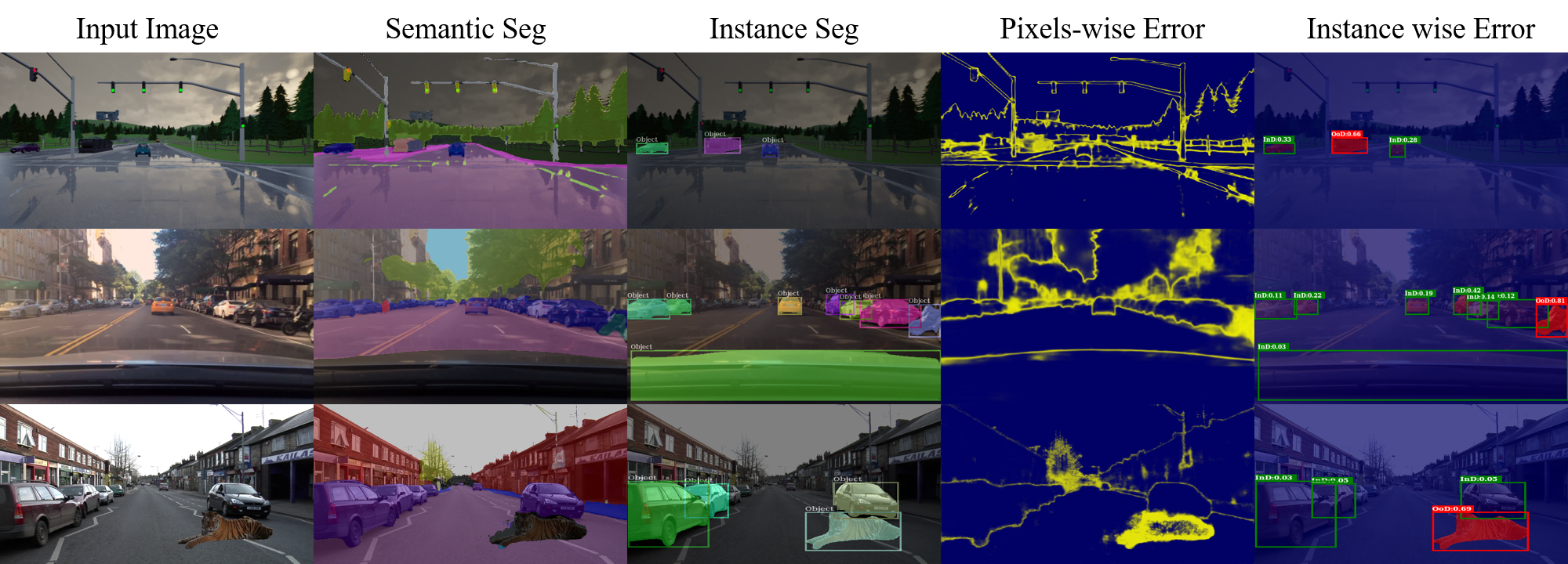}
\caption{\textbf{Qualitative results on the StreetHazards (top row), BDD Anomaly (mid) and CamVid (bot).} From left to right, the input image; the semantic segmentation; the instance segmentation; the pixel-wise error and the instance-wise error. We can see that our method is able to detect numerous objects and disentangle in-distribution objects ($a_{i}<.5$ in green) from out-of-distribution objects ($a_{i}>.5$ in red).}
\label{fig:qualitative_results}
\vspace{-0.5cm}
\end{figure*}

We assess experimentally the effectiveness of our observer network coupled with an class-agnostic instance detector and compare it against several baselines.

\subsection{Datasets \& Metrics}
We conducts experiments on the \textbf{CamVid OOD} \cite{besnier2021learning}, \textbf{StreetHazards} \cite{hendrycks_benchmark_2019} and \textbf{BDD Anomaly} \cite{bdd100k} datasets of urban streets scenes with anomalies in the test set. Anomalies correspond to OOD objects, not seen during training.

To evaluate each method on these datasets, we select four metrics to detect misclassified and out-of-distribution examples: \textbf{fpr95tpr}~\cite{Liang2018}, \textbf{Area Under the Receiver Operating Characteristic curve (AuRoc)}~\cite{hendrycks17baseline}, \textbf{Area under the Precision-Recall Curve (AuPR)}~\cite{hendrycks17baseline} and \textbf{Mean Average Prediction ($mAP_\delta$)}. We compute the latter metric where we discard object smaller than $\delta^2$ pixels.

For each metric, we report the result where an object is considered as well detected if the predicted mask has $IoU > .5$ with the ground truth. We assign to each detected object the anomaly score computed as \autoref{eq:anomalyscore}. We use a Bayesian SegNet \cite{badrinarayanan2015segnet}, \cite{kendall2015bayesian} as the main network for CamVid and a DeepLabv3+~\cite{Chen_2018_ECCV} for BDD Anomaly and StreetHazards. The \obsnet follows the same architecture as the corresponding segmentation network. 

For our instance segmentation module, we select two Mask R-CNN variants~\cite{He_2017_ICCV}: one trained on CityScapes~\cite{Cordts2016Cityscapes}, reported as In-Distribution Detector, and one trained on MS-COCO~\cite{Lin_2014}, reported as Pan-Distribution. We do not leverage the class predicted but only the instance mask. Moreover, we use an additional oracle: we take every connected region of the same class in the annotation as one instance of an object, we report this \textit{detector} as GT-detector.

We compare our method against two other methods. \textbf{MCP}~\cite{hendrycks17baseline}: Maximum Class Prediction; one minus the maximum of the prediction. And \textbf{\mcdropout}~\cite{Gal2016Dropout}: The entropy of the mean softmax prediction with dropout; we use 50 forward passes for all the experiences.

\subsection{Benefit of the instance module}\label{subsection:instance-wise-ablation}
To validate the benefit of the instance detector, we first check that filtering the pixel-wise error map with the instance detector helps for pixel OOD detection, see \autoref{tab_camvid_pixels}. Using an instance detector improves significantly the performance of \obsnet. Moreover, this experiment shows that keeping raw error map is sub-optimal because many pixels with high anomaly score do not correspond to an OOD object but actually belong to the background of the images, whereas they can easily be filtered out by our instance scheme.

\begin{table}
\renewcommand{\figurename}{Table}
\renewcommand{\captionlabelfont}{\bf}
\renewcommand{\captionfont}{\small} 
 \centering
  \begin{tabular}{l@{\hspace{-0.1cm}}c@{\hspace{0.1cm}}c@{\hspace{0.1cm}}c@{\hspace{0.1cm}}c@{\hspace{0.1cm}}} \toprule
    Method                               & fpr95tpr$\downarrow$     & AuPR\_error $\uparrow$   &  AuRoc $\uparrow$  \\ \hline        
    Softmax~\cite{hendrycks17baseline}   & 70.0                     & 11.45                   & 76.7                \\  
    \obsnet                              & 40.5                     & 22.72                   & 87.9                \\ 
    \obsnet + \textit{in-detector}             & 31.3                     & 49.4                    & 92.3                \\   
    \obsnet + \textit{pan-detector}            &  8.7                     & 70.1                    & 97.3                \\   
    \obsnet + \textit{gt-detector}             &  \textbf{1.0}            & \textbf{90.5}           & \textbf{99.7}       \\   
    \bottomrule                                                                                             
  \end{tabular}
  \caption{\textbf{Pixel-wise evaluation on CamVid OOD}. We consider OOD pixels only as the positive class.}
  \label{tab_camvid_pixels}
  \vspace{-0.5cm}
\end{table}

\subsection{Instance-Wise Results}\label{subsection:instance-wise_results}

Here, we compare our methods for object detection on \autoref{tab_camvid_instance}, \autoref{tab_bdd_instance} and \autoref{tab_streethazards_instance}. We can observe that for each dataset the results are quite different. This is due to the scale of the anomalies and the number of them in each the dataset. In \autoref{tab_camvid_instance}, all anomalies are above 64² pixels, which can explain why the metrics drastically improve as we discard smaller detected objects. In \autoref{tab_streethazards_instance}, most of the objects are in fact anomalies, which is why mAP is high, even for the object below 32² pixels. Finally, even if on average \textit{pan-detector} outperforms \textit{in-detector}, this is not always the case in \autoref{tab_bdd_instance}. Indeed, \textit{pan-detector} can detect more objects, and among them smaller in-distribution objects, that can hurt performances. Overall, 
\obsnet outperforms baseline methods, regardless of the detector.

\begin{table}
\renewcommand{\figurename}{Table}
\renewcommand{\captionlabelfont}{\bf}
\renewcommand{\captionfont}{\small} 
 \centering
  \begin{tabular}{l@{\hspace{0.1cm}}l@{\hspace{-0.1cm}}c@{\hspace{0.1cm}}c@{\hspace{0.1cm}}c@{\hspace{0.1cm}}c@{\hspace{0.1cm}}c@{\hspace{0.1cm}}} \toprule
    Det      &  Method                               & $fpr95_{32}\downarrow$ &  $Roc_{32} \uparrow$   & $mAP_0$  & $mAP_{32}$  \\ \hline        
             &  Softmax~\cite{hendrycks17baseline}   & *                       &  56.7                  & 7.7               & 55.2    \\  
    In       &  \mcdropout~\cite{Gal2016Dropout}     & *                       &  58.3                  & 8.4               & 58.5    \\
             &  \obsnet                              & *                       &  \textbf{60.5}         & \textbf{9.9}  & \textbf{63.7}   \\ \hdashline 
             &  Softmax~\cite{hendrycks17baseline}   & 57.2                    &  79.4                  & 4.8               & 62.1    \\  
    Pan      &  \mcdropout~\cite{Gal2016Dropout}     & 52.8                    &  84.6                  & 6.9               & 70.8    \\
             &  \obsnet                              & \textbf{46.4}           &  \textbf{89.6}         & \textbf{11.3}     & \textbf{81.4}     \\ \hdashline
             &  Softmax~\cite{hendrycks17baseline}   & 43.3                    &  80.4                  & 10.8              & 72.1    \\  
    GT       &  \mcdropout~\cite{Gal2016Dropout}     & 32.1                    &  85.5                  & 13.5              & 79.2    \\
             &  \obsnet                              & \textbf{27.2}           &  \textbf{94.3}         & \textbf{22.3}     & \textbf{92.0}   \\ 
    \bottomrule                                                                                             
  \end{tabular}
  \caption{\textbf{Instance-wise evaluation on CamVid OOD}. We consider OOD examples only as the positive class. *not enough OOD objects have been detected by the detector to compute the metrics}
  \label{tab_camvid_instance}
  \vspace{-0.4cm}
\end{table}

\begin{table}
\renewcommand{\figurename}{Table}
\renewcommand{\captionlabelfont}{\bf}
\renewcommand{\captionfont}{\small} 
 \centering
  \begin{tabular}{l@{\hspace{0.1cm}}l@{\hspace{-0.1cm}}c@{\hspace{0.1cm}}c@{\hspace{0.1cm}}c@{\hspace{0.1cm}}c@{\hspace{0.1cm}}c@{\hspace{0.1cm}}} \toprule
    Det   &  Method                                   & $fpr95_{32}\downarrow$ &  $Roc_{32} \uparrow$   & $mAP_0 $  & $mAP_{32}$ \\ \hline        
             &  Softmax~\cite{hendrycks17baseline}   &  *                        &  52.5                  &  9.5           &    13.3            \\  
    In       &  \mcdropout~\cite{Gal2016Dropout}     &  *                        &  52.6                  &  9.5           &    13.3               \\
             &  \obsnet                              &  *                        &  \textbf{55.8}         &  \textbf{9.9}  &    \textbf{16.8}     \\\hdashline      
             &  Softmax~\cite{hendrycks17baseline}   &  *                        &  63.0                  &  5.9           &    16.8            \\  
    Pan      &  \mcdropout~\cite{Gal2016Dropout}     &  *                        &  62.2                  &  6.0           &    17.0            \\
             &  \obsnet                              &  *                        &  \textbf{64.5}         &  \textbf{6.9}  &    \textbf{20.1}          \\ \hdashline      
             &  Softmax~\cite{hendrycks17baseline}   &  65.6                     &  81.9                  &  22.7          &    37.8              \\  
    GT       &  \mcdropout~\cite{Gal2016Dropout}     &  61.9                     &  82.5                  &  23.0          &    39.1              \\
             &  \obsnet                              &  \textbf{53.3}            &  \textbf{86.8}         &  \textbf{27.1} &    \textbf{50.7}  \\ \bottomrule  
  \end{tabular}
  \caption{\textbf{Instance-wise evaluation on BDD Anomaly}. We consider OOD examples only as the positive class.}
  \label{tab_bdd_instance}
  \vspace{-0.2cm}
\end{table}

\begin{table}
\renewcommand{\figurename}{Table}
\renewcommand{\captionlabelfont}{\bf}
\renewcommand{\captionfont}{\small} 
 \centering
  \begin{tabular}{l@{\hspace{0.1cm}}l@{\hspace{-0.1cm}}c@{\hspace{0.1cm}}c@{\hspace{0.1cm}}c@{\hspace{0.1cm}}c@{\hspace{0.1cm}}c@{\hspace{0.1cm}}c@{\hspace{0.1cm}}} \toprule
    Det      &  Method                               & $fpr95_{48}\downarrow$ &  $Roc_{48} \uparrow$   & $mAP_{16} $  & $mAP_{48}$          \\ \hline      
             &  Softmax~\cite{hendrycks17baseline}   & *                    &  \textbf{50.4}         & 80.0             &  \textbf{81.9}    \\
    In       &  \mcdropout~\cite{Gal2016Dropout}     & *                    &  50.3                  & 80.0             &  81.2             \\
             &  \obsnet                              & *                    &  \textbf{50.4}         & \textbf{80.1}    &  \textbf{81.9}    \\\hdashline 
             &  Softmax~\cite{hendrycks17baseline}   & *                    &  53.7                  & 57.6             &  77.7                \\  
    Pan      &  \mcdropout~\cite{Gal2016Dropout}     & *                    &  53.6                  & \textbf{57.7}    &  77.6                \\
             &  \obsnet                              & *                    &  \textbf{54.1}         & 56.9             &  \textbf{77.8}          \\ \hdashline
             &  Softmax~\cite{hendrycks17baseline}   & 80.0                 &  85.2                  & 88.9             &  99.0                          \\  
    GT       &  \mcdropout~\cite{Gal2016Dropout}     & 74.5                 &  86.0                  & 86.9             &  99.0                          \\
             &  \obsnet                              & \textbf{72.6}        &  \textbf{87.5}         & \textbf{89.0}    &  \textbf{99.2}       \\ \bottomrule  
  \end{tabular}
  \caption{\textbf{Instance-wise evaluation on StreetHazards}. We consider OOD examples only as the positive class.}
  \label{tab_streethazards_instance}
  \vspace{-0.4cm}
\end{table}

In \autoref{fig:histogram}, we report the histogram of objects detected by our detector, ranked by our framework. We can well disentangle in-distribution objects as cars, bicycles, or pedestrians, from OOD objects (animals).

We illustrate a few qualitative results in \autoref{fig:qualitative_results}. \obsnet emphasizes OOD objects with higher anomaly scores compared to in-distribution objects. Its predicted error maps are generally clearer and more accurate.

\section{Conclusion}
In this paper, with propose to use an additional, class-agnostic, object detector to filter and aggregate an anomaly score from \obsnet pixel-wise error map. Our strategy helps to better disentangle in from out-of-distribution objects. 

\clearpage
{\small
\bibliographystyle{ieee_fullname}
\bibliography{Sections/citation}

\begin{thebibliography}{10}\itemsep=-1pt

\bibitem{azevedo2020stochastic}
Tiago Azevedo, Ren{\'e} de Jong, Matthew Mattina, and Partha Maji.
\newblock Stochastic-yolo: Efficient probabilistic object detection under
  dataset shifts.
\newblock 2020.

\bibitem{badrinarayanan2015segnet}
Vijay Badrinarayanan, Alex Kendall, and Roberto Cipolla.
\newblock {SegNet}: A deep convolutional encoder-decoder architecture for image
  segmentation.
\newblock {\em IEEE Trans. PAMI}, 2017.

\bibitem{baur2018deep}
Christoph Baur, Benedikt Wiestler, Shadi Albarqouni, and Nassir Navab.
\newblock Deep autoencoding models for unsupervised anomaly segmentation in
  brain {MR} images.
\newblock In {\em MICCAI Workshops}, 2018.

\bibitem{besnier2021triggering}
Victor Besnier, Andrei Bursuc, David Picard, and Alexandre Briot.
\newblock Triggering failures: Out-of-distribution detection by learning from
  local adversarial attacks in semantic segmentation.
\newblock In {\em ICCV}, 2021.

\bibitem{besnier2021learning}
Victor Besnier, David Picard, and Alexandre Briot.
\newblock Learning uncertainty for safety-oriented semantic segmentation in
  autonomous driving.
\newblock In {\em ICIP}, 2021.

\bibitem{bevandic2019simultaneous}
Petra Bevandi{\'c}, Ivan Kre{\v{s}}o, Marin Or{\v{s}}i{\'c}, and Sini{\v{s}}a
  {\v{S}}egvi{\'c}.
\newblock Simultaneous semantic segmentation and outlier detection in presence
  of domain shift.
\newblock In {\em GCPR}, 2019.

\bibitem{Blundell2015}
Charles Blundell, Julien Cornebise, Koray Kavukcuoglu, and Daan Wierstra.
\newblock {Weight uncertainty in neural networks}.
\newblock {\em ICML}, 2015.

\bibitem{Chen_2018_ECCV}
Liang-Chieh Chen, Yukun Zhu, George Papandreou, Florian Schroff, and Hartwig
  Adam.
\newblock Encoder-decoder with atrous separable convolution for semantic image
  segmentation.
\newblock In {\em ECCV}, 2018.

\bibitem{Choi_2019_ICCV}
Jiwoong Choi, Dayoung Chun, Hyun Kim, and Hyuk-Jae Lee.
\newblock Gaussian {YOLO}v3: An accurate and fast object detector using
  localization uncertainty for autonomous driving.
\newblock In {\em ICCV}, 2019.

\bibitem{choi2021active}
Jiwoong Choi, Ismail Elezi, Hyuk-Jae Lee, Clement Farabet, and Jose~M Alvarez.
\newblock Active learning for deep object detection via probabilistic modeling.
\newblock In {\em ICCV}, 2021.

\bibitem{corbiere2019addressing}
Charles Corbi{\`e}re, Nicolas Thome, Avner Bar-Hen, Matthieu Cord, and Patrick
  P{\'e}rez.
\newblock Addressing failure prediction by learning model confidence.
\newblock In {\em NeurIPS}, 2019.

\bibitem{Cordts2016Cityscapes}
Marius Cordts, Mohamed Omran, Sebastian Ramos, Timo Rehfeld, Markus Enzweiler,
  Rodrigo Benenson, Uwe Franke, Stefan Roth, and Bernt Schiele.
\newblock The {C}ityscapes dataset for semantic urban scene understanding.
\newblock In {\em CVPR}, 2016.

\bibitem{devries2018learning}
Terrance DeVries and Graham~W Taylor.
\newblock Learning confidence for out-of-distribution detection in neural
  networks.
\newblock {\em arXiv}, 2018.

\bibitem{du2022stud}
Xuefeng Du, Xin Wang, Gabriel Gozum, and Yixuan Li.
\newblock Unknown-aware object detection: Learning what you don’t know from
  videos in the wild.
\newblock In {\em CVPR}, 2022.

\bibitem{du2022towards}
Xuefeng Du, Zhaoning Wang, Mu Cai, and Sharon Li.
\newblock Towards unknown-aware learning with virtual outlier synthesis.
\newblock In {\em ICLR}, 2022.

\bibitem{durasov2021masksembles}
Nikita Durasov, Timur Bagautdinov, Pierre Baque, and Pascal Fua.
\newblock Masksembles for uncertainty estimation.
\newblock In {\em CVPR}, 2021.

\bibitem{dusenberry2020efficient}
Michael~W. Dusenberry, Ghassen Jerfel, Yeming Wen, Yian Ma, Jasper Snoek,
  Katherine Heller, Balaji Lakshminarayanan, and Dustin Tran.
\newblock Efficient and scalable bayesian neural nets with rank-1 factors.
\newblock In {\em ICML}, 2020.

\bibitem{franchi2020encoding}
Gianni Franchi, Andrei Bursuc, Emanuel Aldea, S{\'e}verine Dubuisson, and
  Isabelle Bloch.
\newblock Encoding the latent posterior of bayesian neural networks for
  uncertainty quantification.
\newblock {\em arXiv}, 2020.

\bibitem{franchi2019tradi}
Gianni Franchi, Andrei Bursuc, Emanuel Aldea, S{\'e}verine Dubuisson, and
  Isabelle Bloch.
\newblock {TRADI}: Tracking deep neural network weight distributions.
\newblock In {\em ECCV}, 2020.

\bibitem{Gal2016}
Yarin Gal.
\newblock {Uncertainty in Deep Learning}.
\newblock {\em PhD}, 2016.

\bibitem{Gal2016Dropout}
Yarin Gal and Zoubin Ghahramani.
\newblock Dropout as a bayesian approximation: Representing model uncertainty
  in deep learning.
\newblock In {\em ICML}, 2016.

\bibitem{Gasperini_2022}
Stefano Gasperini, Jan Haug, Mohammad-Ali~Nikouei Mahani, Alvaro Marcos-Ramiro,
  Nassir Navab, Benjamin Busam, and Federico Tombari.
\newblock Certainnet: Sampling-free uncertainty estimation for object
  detection.
\newblock {\em RAL}, 2022.

\bibitem{guo_2017}
Chuan Guo, Geoff Pleiss, Yu Sun, and Kilian~Q. Weinberger.
\newblock On calibration of modern neural networks.
\newblock {\em ICML}, 2017.

\bibitem{Harakeh2021EstimatingAE}
Ali Harakeh and Steven~L. Waslander.
\newblock Estimating and evaluating regression predictive uncertainty in deep
  object detectors.
\newblock In {\em ICLR}, 2021.

\bibitem{haussmann2020scalable}
Elmar Haussmann, Michele Fenzi, Kashyap Chitta, Jan Ivanecky, Hanson Xu, Donna
  Roy, Akshita Mittel, Nicolas Koumchatzky, Clement Farabet, and Jose~M
  Alvarez.
\newblock Scalable active learning for object detection.
\newblock In {\em IV}, 2020.

\bibitem{He_2017_ICCV}
Kaiming He, Georgia Gkioxari, Piotr Dollar, and Ross Girshick.
\newblock Mask {R-CNN}.
\newblock In {\em ICCV}, 2017.

\bibitem{hecker2018failure}
Simon Hecker, Dengxin Dai, and Luc Van~Gool.
\newblock Failure prediction for autonomous driving.
\newblock In {\em IV}, 2018.

\bibitem{hein2019relu}
Matthias Hein, Maksym Andriushchenko, and Julian Bitterwolf.
\newblock Why relu networks yield high-confidence predictions far away from the
  training data and how to mitigate the problem.
\newblock In {\em CVPR}, 2019.

\bibitem{hendrycks_benchmark_2019}
Dan Hendrycks, Steven Basart, Mantas Mazeika, Mohammadreza Mostajabi, Jacob
  Steinhardt, and Dawn Song.
\newblock A benchmark for anomaly segmentation.
\newblock {\em ArXiv}, 2019.

\bibitem{hendrycks17baseline}
Dan Hendrycks and Kevin Gimpel.
\newblock A baseline for detecting misclassified and out-of-distribution
  examples in neural networks.
\newblock In {\em ICLR}, 2017.

\bibitem{hendrycks2018deep}
Dan Hendrycks, Mantas Mazeika, and Thomas Dietterich.
\newblock Deep anomaly detection with outlier exposure.
\newblock In {\em ICLR}, 2018.

\bibitem{kendall2015bayesian}
Alex Kendall, Vijay Badrinarayanan, , and Roberto Cipolla.
\newblock Bayesian {SegNet}: Model uncertainty in deep convolutional
  encoder-decoder architectures for scene understanding.
\newblock {\em arXiv}, 2015.

\bibitem{KendallGal_2017}
Alex Kendall and Yarin Gal.
\newblock What uncertainties do we need in bayesian deep learning for computer
  vision?
\newblock In {\em NeurIPS}, 2017.

\bibitem{Lakshminarayanan_2017}
Balaji Lakshminarayanan, Alexander Pritzel, and Charles Blundell.
\newblock Simple and scalable predictive uncertainty estimation using deep
  ensembles.
\newblock In {\em NeurIPS}, 2017.

\bibitem{lee2018simple}
Kimin Lee, Kibok Lee, Honglak Lee, and Jinwoo Shin.
\newblock A simple unified framework for detecting out-of-distribution samples
  and adversarial attacks.
\newblock In {\em NeurIPS}, 2018.

\bibitem{Liang2018}
Shiyu Liang, R. Srikant, and Yixuan Li.
\newblock Enhancing the reliability of out-of-distribution image detection in
  neural networks.
\newblock In {\em ICLR}, 2018.

\bibitem{Lin_2014}
Tsung-Yi Lin, Michael Maire, Serge Belongie, James Hays, Pietro Perona, Deva
  Ramanan, Piotr Doll{\'a}r, and C.~Lawrence Zitnick.
\newblock Microsoft {COCO}: Common objects in context.
\newblock In {\em ECCV}, 2014.

\bibitem{lis2019detecting}
Krzysztof Lis, Krishna Nakka, Pascal Fua, and Mathieu Salzmann.
\newblock Detecting the unexpected via image resynthesis.
\newblock In {\em ICCV}, 2019.

\bibitem{liu2020simpleSNGP}
Jeremiah~Zhe Liu, Zi Lin, Shreyas Padhy, Dustin Tran, Tania Bedrax-Weiss, and
  Balaji Lakshminarayanan.
\newblock Simple and principled uncertainty estimation with deterministic deep
  learning via distance awareness.
\newblock In {\em NeurIPS}, 2020.

\bibitem{liu2016ssd}
Wei Liu, Dragomir Anguelov, Dumitru Erhan, Christian Szegedy, Scott Reed,
  Cheng-Yang Fu, and Alexander~C Berg.
\newblock Ssd: Single shot multibox detector.
\newblock In {\em ECCV}, 2016.

\bibitem{maddox2019simple}
Wesley~J Maddox, Pavel Izmailov, Timur Garipov, Dmitry~P Vetrov, and
  Andrew~Gordon Wilson.
\newblock A simple baseline for bayesian uncertainty in deep learning.
\newblock In {\em NeurIPS}, 2019.

\bibitem{malinin2018}
Andrey Malinin and Mark Gales.
\newblock Predictive uncertainty estimation via prior networks.
\newblock In {\em NeurIPS}, 2018.

\bibitem{mehrtash2020pep}
Alireza Mehrtash, Purang Abolmaesumi, Polina Golland, Tina Kapur, Demian
  Wassermann, and William~M Wells~III.
\newblock Pep: Parameter ensembling by perturbation.
\newblock In {\em NeurIPS}, 2020.

\bibitem{miller2019evaluating}
Dimity Miller, Feras Dayoub, Michael Milford, and Niko S{\"u}nderhauf.
\newblock Evaluating merging strategies for sampling-based uncertainty
  techniques in object detection.
\newblock In {\em ICRA}, 2019.

\bibitem{naeini2015obtaining}
Mahdi~Pakdaman Naeini, Gregory Cooper, and Milos Hauskrecht.
\newblock Obtaining well calibrated probabilities using bayesian binning.
\newblock In {\em AAAI}, 2015.

\bibitem{naeini2016binary}
Mahdi~Pakdaman Naeini and Gregory~F Cooper.
\newblock Binary classifier calibration using an ensemble of near isotonic
  regression models.
\newblock In {\em KDD}, 2016.

\bibitem{neal2012bayesian}
Radford~M Neal.
\newblock Bayesian learning for neural networks.
\newblock 2012.

\bibitem{nguyen2015deep}
A. {Nguyen}, J. {Yosinski}, and J. {Clune}.
\newblock Deep neural networks are easily fooled: High confidence predictions
  for unrecognizable images.
\newblock In {\em CVPR}, 2015.

\bibitem{platt1999probabilistic}
John Platt et~al.
\newblock Probabilistic outputs for support vector machines and comparisons to
  regularized likelihood methods.
\newblock {\em ALMC}, 1999.

\bibitem{postels2019sampling}
Janis Postels, Francesco Ferroni, Huseyin Coskun, Nassir Navab, and Federico
  Tombari.
\newblock Sampling-free epistemic uncertainty estimation using approximated
  variance propagation.
\newblock In {\em ICCV}, 2019.

\bibitem{rahman2020online}
Quazi~Marufur Rahman, Niko S{\"u}nderhauf, and Feras Dayoub.
\newblock Online monitoring of object detection performance post-deployment.
\newblock In {\em IROS}, 2020.

\bibitem{ren2015faster}
Shaoqing Ren, Kaiming He, Ross Girshick, and Jian Sun.
\newblock Faster r-cnn: Towards real-time object detection with region proposal
  networks.
\newblock In {\em NeurIPS}, 2015.

\bibitem{Samson_2019_ICCV}
Laurens Samson, Nanne van Noord, Olaf Booij, Michael Hofmann, Efstratios
  Gavves, and Mohsen Ghafoorian.
\newblock I bet you are wrong: Gambling adversarial networks for structured
  semantic segmentation.
\newblock In {\em ICCV Workshops}, 2019.

\bibitem{schlegl2017unsupervised}
Thomas Schlegl, Philipp Seeb{\"o}ck, Sebastian~M Waldstein, Ursula
  Schmidt-Erfurth, and Georg Langs.
\newblock Unsupervised anomaly detection with generative adversarial networks
  to guide marker discovery.
\newblock In {\em IPMI}, 2017.

\bibitem{srivastava2014dropout}
Nitish Srivastava, Geoffrey Hinton, Alex Krizhevsky, Ilya Sutskever, and Ruslan
  Salakhutdinov.
\newblock Dropout: A simple way to prevent neural networks from overfitting.
\newblock {\em JMLR}, 2014.

\bibitem{van2020uncertainty}
Joost Van~Amersfoort, Lewis Smith, Yee~Whye Teh, and Yarin Gal.
\newblock Uncertainty estimation using a single deep deterministic neural
  network.
\newblock In {\em ICML}, 2020.

\bibitem{wen2020batchensemble}
Yeming Wen, Dustin Tran, and Jimmy Ba.
\newblock Batchensemble: an alternative approach to efficient ensemble and
  lifelong learning.
\newblock In {\em ICLR}, 2020.

\bibitem{xia2020synthesize}
Yingda Xia, Yi Zhang, Fengze Liu, Wei Shen, and Alan Yuille.
\newblock Synthesize then compare: Detecting failures and anomalies for
  semantic segmentation.
\newblock In {\em ECCV}, 2020.

\bibitem{yoo2019learning}
Donggeun Yoo and In~So Kweon.
\newblock Learning loss for active learning.
\newblock In {\em CVPR}, 2019.

\bibitem{bdd100k}
Fisher Yu, Haofeng Chen, Xin Wang, Wenqi Xian, Yingying Chen, Fangchen Liu,
  Vashisht Madhavan, and Trevor Darrell.
\newblock {BDD100K}: A diverse driving dataset for heterogeneous multitask
  learning.
\newblock In {\em CVPR}, 2020.

\bibitem{zadrozny2001obtaining}
Bianca Zadrozny and Charles Elkan.
\newblock Obtaining calibrated probability estimates from decision trees and
  naive bayesian classifiers.
\newblock In {\em ICML}, 2001.

\bibitem{zadrozny2002transforming}
Bianca Zadrozny and Charles Elkan.
\newblock Transforming classifier scores into accurate multiclass probability
  estimates.
\newblock In {\em KDD}, 2002.

\end{thebibliography}
}

\end{document}